%% file: od_prediction/main.tex
\documentclass[11 pt, onecolumn]{article}

\usepackage[margin=1.1in]{geometry}
\input{od_prediction/sections/header}
\usepackage{algorithm}
\usepackage{algpseudocode}
\usepackage{caption}
\usepackage[most]{tcolorbox}
\usepackage[most]{tcolorbox}
\usepackage{enumitem}
\DeclareMathAlphabet{\mathcal}{OMS}{cmsy}{m}{n}

\usepackage{graphicx}
\usepackage{epstopdf}

\begin{document}
\title{\LARGE \bf Standard Applicability Judgment and Cross-jurisdictional
Reasoning: A RAG-based Framework for Medical Device
Compliance}

\author{Yu Han, Aaron Ceross, and Jeroen H.M. Bergmann
\thanks{Yu Han is with the Institute of Biomedical Engineering, Department of Engineering Science, University of Oxford, Oxford, UK. Aaron Ceross is with Birmingham Law School, University of Birmingham, Birmingham, UK. Emails: yu.han@ox.ac.uk, a.ceross@bham.ac.uk, jeroen.bergmann@eng.ox.ac.uk}%
}
\newcommand*{\QEDA}{\hfill\ensuremath{\blacksquare}}%

\maketitle

\begin{abstract}
Identifying the appropriate regulatory standard applicability remains a critical yet understudied challenge in medical device compliance, frequently necessitating expert interpretation of fragmented and heterogeneous documentation across different jurisdictions. To address this challenge, we introduce a modular AI system that leverages a retrieval-augmented generation (RAG) pipeline to automate standard applicability determination. Given a free-text device description, our system retrieves candidate standards from a curated corpus and uses large language models to infer jurisdiction-specific applicability—classified as Mandatory, Recommended, or Not Applicable—with traceable justifications. We construct an international benchmark dataset of medical device descriptions with expert-annotated standard mappings, and evaluate our system against retrieval-only, zero-shot, and rule-based baselines. The proposed approach attains a classification accuracy of 73\% and a Top-5 retrieval recall of 87\%, demonstrating its effectiveness in identifying relevant regulatory standards. We introduce the first end-to-end system for standard applicability reasoning, enabling scalable and interpretable AI-supported regulatory science. Notably, our region-aware RAG agent performs cross-jurisdictional reasoning between Chinese and U.S. standards, supporting conflict resolution and applicability justification across regulatory frameworks.
\end{abstract}


\input{od_prediction/sections/introduction}
\input{od_prediction/sections/model}

\input{od_prediction/sections/experiments}
\input{od_prediction/sections/conclusion}

\bibliographystyle{IEEEtran}
\bibliography{od_prediction/od_prediction} 
\end{document}

%% file: od_prediction/sections/header.tex
\usepackage{amsmath}
\usepackage{bm}	
\usepackage{graphicx}

\usepackage{graphicx}
\usepackage{multirow}
\usepackage{mdwlist}

\usepackage{threeparttable}

\usepackage{amsfonts}
\usepackage{amsmath}
\usepackage{hyperref}
\usepackage{bbm}
\usepackage{float}
\usepackage{caption}
\usepackage{subcaption}
\usepackage{xcolor}
\usepackage{mathtools}
\usepackage{tikz}

\usetikzlibrary{shapes.geometric,arrows}
\usepackage{enumerate}
\usepackage{multicol}
\usepackage{stackrel}

\usepackage{subcaption}
\usepackage{amssymb}
\usepackage[utf8]{inputenc}
\usepackage{booktabs,lipsum}
\usepackage{hyperref}

%% file: od_prediction/sections/introduction.tex
\section{Introduction}
Regulatory standard applicability assessment, is a crucial step where determining which safety or performance standards apply to a given medical device—is a critical yet largely heuristic task in compliance workflows. Modern regulatory systems involve thousands of complex, domain-specific standards that vary significantly across jurisdictions. For medical device manufacturers, determining the applicability of these standards is a high-stakes, labor-intensive task that often requires deep expertise and manual reasoning \cite{han2024more}.

Regulatory standards for medical devices differ substantially across national jurisdictions, reflecting variations in legal frameworks, healthcare infrastructure, and risk tolerance. While the International Medical Device Regulators Forum (IMDRF) and initiatives such as the Global Harmonization Task Force (GHTF) aim to promote convergence, significant discrepancies persist between regulatory authorities like the U.S. Food and Drug Administration (FDA), the European Medicines Agency (EMA), and China's NMPA \cite{li2021fdaai, chalkidis2021lexglue, ema_guidance2024, han2024regulatory, han2024transforming}. These divergences manifest in definitions of device classes, clinical evidence requirements, software validation procedures, and post-market surveillance obligations \cite{bommasani2022opportunities, ogilvie2022chatbots, crudeli2020quality}. For global manufacturers holding a single Technical Requirements Document (TRD), this regulatory fragmentation imposes compliance friction, as a TRD approved in one country may fail to meet the expectations of another—requiring duplication of effort, additional documentation, or reengineering of safety claims \cite{zhang2022ai, bran2023chemcrow}. This lack of interoperability not only delays time-to-market but increases the burden on regulatory affairs teams to maintain multiple localized dossiers \cite{castelvecchi2023opensource, anand2023query}.

Moreover, LLM-based tools that assist with regulatory mapping or justification must account for jurisdiction-specific clauses, legal language, and precedence, as assumptions valid under FDA’s CFR §820 may not apply under EU MDR Annex IX or China's YY/T series \cite{lewis2020retrieval, wen2023grove}. Addressing this complexity requires both robust retrieval-augmented reasoning systems that are context-aware and global policy efforts to align technical expectations, especially as device software and AI modules grow in importance \cite{zhang2023retrieval, hu2021lora, grusky2023rogue}.

Recent advances in large language models (LLMs) and retrieval-augmented generation (RAG) architectures offer new possibilities for automated, interpretable decision support in high-stakes domains \cite{lewis2020retrieval}. While LLMs can reason over unstructured input and generate natural language justifications, their performance improves significantly when paired with retrieval systems that inject domain-specific knowledge into the inference process \cite{izacard2021leveraging}. This motivates our proposal of a modular intelligent agent that performs standard applicability judgment by combining semantic retrieval over a structured regulatory knowledge base with LLM-driven reasoning and explanation generation.

Unlike generic knowledge-based QA systems, standard applicability judgment for medical devices involves high-stakes, jurisdiction-dependent decisions that are not explicitly codified in the standard text. Applicability cannot be determined by simple keyword matching or retrieval alone, as it often depends on nuanced interpretation of both the device functionality and regulatory intent. AI systems face three challenges in this domain: (1) navigating large unstructured standard corpora, (2) aligning recommendations with domain knowledge and legal language, and (3) providing traceable and explainable decisions to support regulatory auditability.

We introduce a region-aware, end-to-end system for regulatory standard applicability judgment in medical device compliance. Distinct from prior RAG-based approaches in legal and biomedical domains, our framework targets the structural, jurisdictional, and traceability challenges unique to regulatory science. To our knowledge, this is the first system that operationalizes regulatory standard applicability as a retrieval-augmented inference task using large language models, with structured, region-aware explanations. Our key contributions include:

\begin{itemize}
    \item We construct a multilingual regulatory knowledge base comprising over 3,900 Chinese and U.S. medical device standards, each annotated with structured metadata including scope, applicability conditions, technical fields, and limitations. This required the integration, parsing, and manual verification of heterogeneous data sources across jurisdictions, representing a substantial effort in large-scale data curation for regulatory AI.

    \item We release a manually annotated benchmark of 105 product descriptions and their applicability mappings labeled as Mandatory, Recommended, or Not Applicable;

    \item We implement a region-aware reasoning pipeline that enables cross-jurisdictional conflict detection, clause-level gap analysis, and structured JSON outputs for traceable regulatory audit;

    \item We evaluate our system with strict quantitative metrics, demonstrating superior performance over retrieval-only and zero-shot baselines.
\end{itemize}

\section{Related Work}
Recent advances in foundation models have dramatically reshaped the landscape of AI applications in high-stakes domains, including healthcare and regulation. Bommasani et al. systematically examined the transformative potential and associated risks of foundation models, highlighting their capacity for generalization across tasks as well as the need for rigorous alignment, traceability, and domain-specific grounding in deployment-critical environments such as law and medicine \cite{bommasani2022opportunities}. In the context of medical AI, trust, interpretability, and human-centered design have become central pillars for clinical integration. AI systems must augment—not replace—human decision-making in healthcare, supporting transparent, evidence-based workflows while preserving clinician judgment and patient agency \cite{topol2019deep}. These perspectives underscore the need for AI systems in regulatory science to be both scalable and trustworthy, with built-in mechanisms for explanation, domain alignment, and human oversight. Our work situates itself at this intersection, proposing a retrieval-augmented foundation model architecture that directly addresses these emerging requirements in regulatory applicability judgment for medical devices.

There is also increasing interest in the application of artificial intelligence to regulatory processes and compliance decision-making. Notable efforts have included the use of machine learning for adverse event detection, document classification in clinical trials, and ontology-driven rule extraction for medical device labeling \cite{zhang2022ai, li2021fdaai}. However, few systems to date have addressed the specific problem of standard applicability assessment, which requires both domain-specific retrieval and explainable reasoning.

\subsection{AI for Regulatory and Compliance Automation}

The integration of artificial intelligence (AI) into regulatory processes has garnered significant attention, particularly in the medical device industry. AI-driven solutions have been employed to enhance various aspects of regulatory compliance, including adverse event detection, document classification, and risk management. For instance, AI-powered platforms have been developed to streamline the compilation and analysis of technical documentation required for regulatory submissions, thereby reducing errors and expediting approvals \cite{turn0search8}. Moreover, the FDA has recognized the transformative potential of AI in medical devices, leading to the development of frameworks and guidelines to accommodate AI/ML-based software \cite{turn0search4}.

Despite these advancements, the specific challenge of assessing the applicability of regulatory standards to medical devices remains underexplored. This task necessitates not only the retrieval of relevant standards but also the interpretation and reasoning over complex regulatory texts to determine their pertinence to specific devices. The complexity is further compounded by the hierarchical and interrelated nature of regulatory documents, which often require nuanced understanding and contextual analysis. Our work addresses this gap by formalizing standard applicability assessment as a retrieval-augmented semantic inference problem, leveraging AI to navigate and interpret the intricate landscape of medical device regulations.

\subsection{Legal and Technical Document Understanding}

Natural Language Processing (NLP) techniques have been increasingly applied to the legal and technical domains to facilitate the analysis and comprehension of complex documents. In the legal field, NLP has been utilized for tasks such as contract analysis, legislation question answering, and regulatory policy comparison \cite{chalkidis2021lexglue, zhong2020jec}. These applications often involve transformer-based models fine-tuned on legal corpora to perform classification or retrieval tasks.

In the context of medical device regulations, NLP has been employed to extract and interpret information from regulatory documents, aiding in compliance and risk assessment. For example, NLP techniques have been used to analyze clinical evaluation reports, enabling manufacturers to stay abreast of reporting requirements and streamline the documentation process \cite{turn0search17}. Additionally, NLP has facilitated the aggregation of field safety notices, enhancing the ability to monitor and respond to safety concerns \cite{turn0search21}.

However, the application of NLP to the specific task of determining the applicability of regulatory standards to medical devices is limited. This task requires a more sophisticated approach that combines information retrieval with semantic reasoning to assess the relevance of standards to specific device contexts. Our research contributes to this area by developing a system that integrates retrieval-augmented generation with domain-specific knowledge bases to support standard applicability assessment.

\subsection{Retrieval-Augmented Generation}

Retrieval-Augmented Generation (RAG) has emerged as a powerful paradigm that combines the generalization capabilities of large language models with domain-specific grounding through document retrieval \cite{lewis2020retrieval, izacard2021leveraging}. In this framework, relevant documents are retrieved from a knowledge base and used to inform the generation of responses, enhancing the accuracy and relevance of the output.

RAG has been applied to various tasks, including open-domain question answering, fact verification, and summarization. In the legal domain, RAG has been utilized to improve the performance of AI systems in tasks requiring nuanced understanding and interpretation of legal texts \cite{turn0search6}. By grounding responses in retrieved documents, RAG helps mitigate issues such as hallucinations and enhances the explainability of AI-generated outputs \cite{turn0news43}.

In the context of regulatory compliance, RAG offers a promising approach to address the challenges associated with interpreting complex and hierarchical regulatory documents. By retrieving pertinent standards and integrating them into the reasoning process, RAG-based systems can provide more accurate and contextually relevant assessments of standard applicability. Our work leverages this paradigm by constructing a structured knowledge base of medical device standards and integrating it into a retrieval-augmented reasoning pipeline to support regulatory decision-making.

\subsection{LLM-Based Decision-Making Agents}

The advent of large language models (LLMs) has paved the way for the development of autonomous agents capable of performing complex decision-making tasks. These agents typically comprise modular architectures that include perception, retrieval, reasoning, and action components, enabling them to process information, draw inferences, and execute decisions \cite{bommasani2022opportunities}.

In high-stakes domains such as healthcare and law, LLM-based agents have been explored for applications including clinical decision support, legal analysis, and policy evaluation. For instance, LLMs have been employed to assist in the interpretation of medical literature, support diagnostic decision-making, and analyze legal documents for compliance purposes \cite{turn0search15}.

The deployment of LLM-based agents in regulatory science necessitates careful consideration of factors such as transparency, accountability, and alignment with domain-specific knowledge. To address these concerns, our work implements a modular decision-making agent that integrates retrieval-augmented generation with domain-specific knowledge bases, enabling the agent to perform multi-stage standard applicability analysis with explainable reasoning. This approach aligns with the principles outlined by regulatory authorities for the responsible use of AI in regulatory activities \cite{turn0search31}.

This study is a novel attempt to address regulatory standard applicability as a retrieval-augmented inference task using a large language model agent in the medical device industry. Our approach bridges the domains of legal and technical NLP, regulatory knowledge engineering, and interpretable decision support, providing a novel paradigm for trustworthy AI in regulatory science. By integrating structured knowledge bases with advanced AI techniques, our system offers a scalable and transparent solution for assessing the applicability of regulatory standards to medical devices, thereby enhancing compliance and supporting innovation in the industry.

%% file: od_prediction/sections/model.tex
\section{Methods}

\begin{figure}[H]
   \centering
\includegraphics[width=0.99\textwidth]{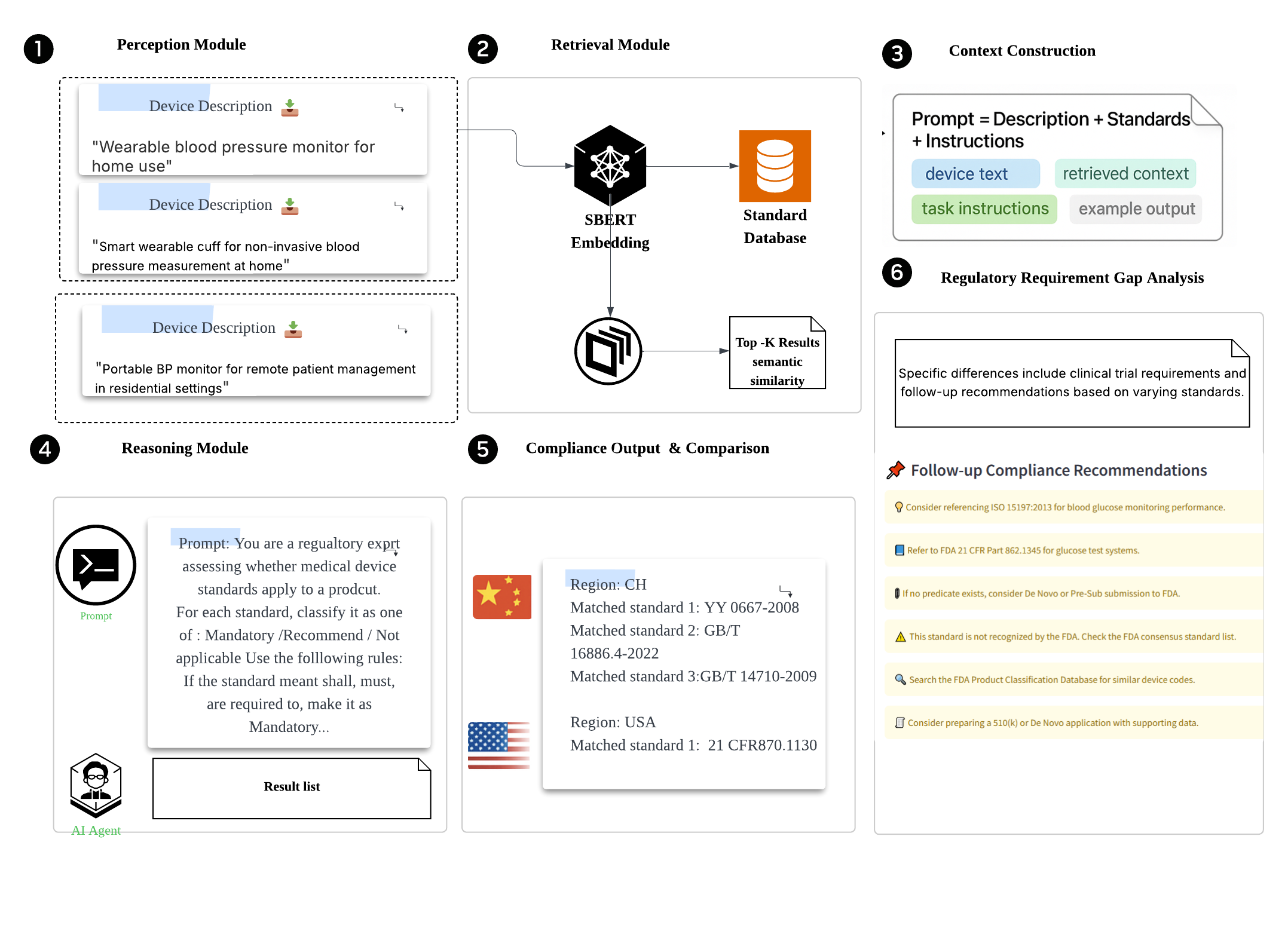}
    \caption{Perception, Retrieval, Context Construction, Reasoning, Compliance Output, and Regulatory Requirement Gap Analysis.}
    \label{fig:overview}
\end{figure}

The designed system comprises six modules as shown in Figure~\ref{fig:overview}), processing a free-text medical device description $x$ against a structured corpus of standards $S = \{s_1, s_2, \dots, s_n\}$. Each standard $s_i$ includes metadata such as an identifier, region, and scope. The model computes a ranked set of applicability tuples $\langle s_i, y_i, e_i \rangle$, where $y_i \in \{\text{Mandatory}, \text{Recommended}, \text{Not Applicable}\}$ is the predicted applicability label and $e_i$ is the corresponding justification.

\[
f(x, S) \rightarrow \left\{ \langle s_i, y_i, e_i \rangle \right\}
\]

where:
\begin{itemize}
    \item $y_i \in \{\text{Mandatory}, \text{Recommended}, \text{Not Applicable} \}$ indicates the applicability classification.
    \item $e_i$ is a plain-language justification for the decision.
\end{itemize}

\subsection{Problem Formulation}

We define the task of \textit{regulatory standard applicability judgment} as a retrieval-augmented semantic classification problem.

\noindent
Given:
\begin{itemize}
    \item A device description $x \in \mathcal{X}$;
    \item A regulatory corpus $S = \{s_1, ..., s_n\}$, where each standard $s_i$ contains a textual description $s_i^{\text{text}}$ and structured metadata $\{r_i, t_i, c_i, \dots\}$.
\end{itemize}

\noindent
Our goal is to construct a function:
\[
f(x, S) \mapsto \{(s_i, y_i, e_i)\ |\ s_i \in \text{Top-}k(S, x)\}
\]
where:
\begin{itemize}
    \item $y_i \in \{\text{Mandatory}, \text{Recommended}, \text{Not Applicable}\}$ is the applicability classification label;
    \item $e_i \in \mathcal{Y}_{\text{text}}$ is the natural language justification associated with each classification.
\end{itemize}

\noindent
We first retrieve Top-$k$ candidate standards using dense semantic retrieval. Let:
\[
\mathbf{v}_x = \text{SBERT}(x), \quad \mathbf{v}_{s_i} = \text{SBERT}(s_i^{\text{text}})
\]
Then the Top-$k$ relevant standards are selected by cosine similarity:
\[
\text{Top-}k(S, x) = \underset{s_i \in S}{\text{arg top-}k} \, \cos(\mathbf{v}_x, \mathbf{v}_{s_i})
\]

\noindent
Next, we apply a large language model $\mathcal{L}$ (e.g., GPT-4) to infer the applicability label and generate an explanation:
\[
(y_i, e_i) = \mathcal{L}(x, s_i^{\text{text}}, \text{prompt template})
\]

\noindent
This produces a structured, explainable applicability judgment for each retrieved standard.

\subsection{Regulatory Corpus Construction and Encoding}

To support multilingual and cross-jurisdictional standard applicability reasoning, we constructed a structured regulatory corpus (\texttt{allreg.json}) by aggregating official medical device standards from China and the United States.

\textbf{Chinese corpus} entries were manually extracted from the \href{https://www.nmpa.gov.cn/zhuanti/cxylqx/ylqxxgmlhz/20210531160922192.html?type=pc}{NMPA standard catalog}, covering YY (industry) and GB (national) standards. The raw TXT files were parsed and normalized into structured entries. \textbf{U.S. corpus} data were sourced from the \href{https://www.accessdata.fda.gov/scripts/cdrh/cfdocs/cfstandards/results.cfm}{FDA Recognized Consensus Standards database}, including 21 CFR parts, ANSI, AAMI, ASTM standards, and CDRH device classifications. Manual parsing scripts converted HTML tables into JSON format entries.

Each standard record contains the following fields:

\begin{tcolorbox}[title=Key Fields in \texttt{allreg.json}, colback=gray!5, colframe=gray!60!black, sharp corners=south]
\begin{description}[style=nextline]
  \item[\texttt{title\_cn / title\_en}] Standard name in Chinese and English
  \item[\texttt{scope\_cn / scope\_en}] Scope of applicability
  \item[\texttt{usage\_condition / limitation}] Explicit use cases and exclusions
  \item[\texttt{source\_text}] Extracted regulatory text or description
  \item[\texttt{region}] Jurisdiction (e.g., ``CN'' or ``US'')
  \item[\texttt{status / dates / organization}] Status, publication/effective dates, issuing agency
  \item[\texttt{technical\_field}] Domain tags (e.g., ``Orthopedic Devices'', ``Joint Replacement'')
  \item[\texttt{tags}] Application-specific keywords for semantic filtering
\end{description}
\end{tcolorbox}

An example entry for a total elbow joint prosthesis includes both YY/T 0606.4-2015 (CN) and 21 CFR 888.3150 (US), with detailed cross-lingual fields describing scope, conditions of use, and limitations. These structured fields enable downstream semantic encoding, standard retrieval, and jurisdictional comparison. The final JSON dataset comprises over 3,900 entries.

The Perception module transforms an unstructured free-text description of a medical device into a dense semantic representation suitable for similarity-based retrieval. We employ Sentence-BERT~\cite{reimers2019sentence}, a transformer-based sentence encoder fine-tuned for semantic textual similarity (STS) tasks. Specifically, we use the \texttt{paraphrase-MiniLM-L6-v2} variant, which is based on a 6-layer DistilBERT architecture with a dimension size of $d=384$, offering a favorable trade-off between embedding fidelity and computational efficiency.

Formally, let $x$ denote the input device description. We obtain its embedding $v_x \in \mathbb{R}^d$ by applying:

\[
v_x = \texttt{SBERT}_{\texttt{MiniLM}}(x)
\]

This vector encodes latent semantics across lexical, functional, and clinical aspects such as anatomical targeting (e.g., "respiratory support"), modality (e.g., "non-invasive"), and usage context (e.g., "home monitoring" or "ICU use"). We apply minimal pre-processing beyond Unicode normalization and punctuation standardization to preserve medical-specific lexical variations.

The output vector $v_x$ serves as input to the Retrieval module, which computes similarity with regulatory standard embeddings. The encoding process is performed once per query and cached for batch evaluation during benchmarking or inference.

\subsection{Semantic Retrieval and Reranking}

We employed \texttt{Sentence-BERT} (\texttt{all-mpnet-base-v2}) to encode each standard entry into a 768-dimensional semantic vector. FAISS (\texttt{IndexFlatL2}) was used to construct the retrieval index. To enhance retrieval precision, we implemented the \texttt{SegmentedRetrieverWithRerank} class, which encodes concatenated segments (title, scope, usage condition, limitation) using \texttt{BAAI/bge-m3}. For each device query, the retriever computes top-$k$ candidates via cosine similarity, then reranks using \texttt{cross-encoder/ms-marco-MiniLM-L-6-v2}.

In addition, we implemented a hybrid keyword matching module with domain-aware expansion. A synonym dictionary was used to expand queries (e.g., ``\textit{glucose}'' \textrightarrow ``\textit{sugar, CGM, blood sugar}'') before computing token overlap and applying strong ID match bonuses. This combination improves recall and interpretability.

The Retrieval module identifies regulatory standards relevant to a medical device description by performing semantic similarity matching across a heterogeneous corpus of international standards. Each standard entry $s_i$ is represented as a structured JSON object with fields including: \texttt{id} (unique identifier), \texttt{name} (title), \texttt{scope} (free-text applicability description), \texttt{region} (e.g., CN for China, US for United States), and \texttt{status} (e.g., active, withdrawn).

To ensure consistency across regions, we encode the \texttt{scope} text of each standard into a dense semantic vector $v_{s_i} \in \mathbb{R}^d$ using a unified Sentence-BERT encoder (\texttt{paraphrase-MiniLM-L6-v2}, $d=384$). The same encoder is used to transform the free-text medical device description into a query vector $v_x$.

Cosine similarity is computed between the device vector and each standard vector:
\[
\text{score}(x, s_i) = \frac{v_x \cdot v_{s_i}}{\|v_x\| \|v_{s_i}\|}
\]

To support efficient retrieval over a large multilingual corpus, we use FAISS~\cite{johnson2019billion} with an inner-product based flat index (\texttt{IndexFlatIP}), which approximates cosine similarity on normalized vectors. The top-$k$ most relevant standards are selected as input to the reasoning module.

Our corpus includes standards from multiple jurisdictions:
\begin{itemize}
  \item \textbf{China}: GB (Guobiao) national standards and YY (industry) standards from the National Medical Products Administration (NMPA).
  \item \textbf{United States}: FDA Code of Federal Regulations (CFR), ANSI, AAMI, and ASTM standards curated from the U.S. FDA Recognized Consensus Standards database.
\end{itemize}

This cross-region retrieval capability enables consistent compliance reasoning across different regulatory regimes, supporting applications in international submissions, market entry assessments, and harmonization analysis.

\subsection{LLM-Based Applicability Classification}

The prompt is designed to support multi-label classification, requiring the LLM to assign each standard to one of the categories: \textit{Mandatory}, \textit{Recommended}, or \textit{Not Applicable}. To facilitate this, the instructions emphasize the identification of regulatory cue words such as "shall", "must", and "should", which frequently indicate applicability in legal and technical documents.

Few-shot examples at the end of the prompt significantly improve output consistency. Each example contains a standard with its applicability label, justification, and clause. The LLM is instructed to output results in structured JSON format using keys such as: \textit{standard\_id}, \textit{applicability}, \textit{justification}

This structured context construction ensures that the LLM’s generation is both syntactically parsable and semantically traceable, enabling downstream validation, auditing, and visualization within compliance workflows.

The top-$k$ retrieved standards are passed into a GPT model via the OpenAI API. Prompts are constructed using the device description, standard metadata, and task instructions emphasizing regulatory cue words (e.g., ``shall'', ``must''). Few-shot examples demonstrate expected JSON outputs with fields: \texttt{standard\_id}, \texttt{applicability}, \texttt{justification}, \texttt{clause}. Temperature is set to 0.3 to ensure stability.

Parsed outputs are post-processed to restore region and clause information and to ensure validity. A fallback pseudo-labeling routine is used during evaluation by mapping gold labels in \texttt{benchmark.csv} for consistency tests.

To enable reliable and domain-aligned reasoning by the large language model (LLM), we construct a structured prompt that embeds relevant context in a controlled and semantically meaningful format. This prompt includes:

\begin{itemize}
    \item The original medical device description, denoted as $x$,
    \item A list of the top-$k$ retrieved standards, each with identifier, region, and scope text,
    \item Explicit task instructions for applicability classification,
\end{itemize}

\subsection{Reasoning Module: Applicability Classification via GPT}

The structured prompt, consisting of the device description $x$, the scopes and metadata of the top-$k$ retrieved standards, and explicit classification instructions, is passed to GPT-3.5~\cite{openai2023gpt4} via the OpenAI Chat Completion API. The model is queried using the \texttt{gpt-4} endpoint, with temperature set to $T=0.3$ to balance deterministic output and linguistic fluency.

The prompt includes task instructions, cue-word rules (e.g., "shall" \textrightarrow
Mandatory), and few-shot JSON examples. These elements guide the LLM to output structured applicability decisions with traceable justifications:

\begin{quote}
\ttfamily
[
  \{\\
  \ \ \ "standard\_id": "YY 0667-2008",\\
  \ \ \ "name": "Medical Oxygen Concentrator",\\
  \ \ \ "applicability": "Mandatory",\\
  \ \ \ "justification": "This standard uses 'shall' and directly addresses the described device.",\\
  \ \ \ "clause": "Section 3.1"\\
  \}
]
\end{quote}

The API returns a single message response, from which the structured content is extracted using Python's \texttt{json.loads()} function. A post-processing step enriches each record by reinserting original metadata (\texttt{region}, \texttt{name}, \texttt{clause}) from the corresponding retrieved standard, ensuring alignment between the retrieved evidence and the generated inference.

We implement exception handling to capture failures in JSON decoding, and optionally retry or surface the raw model output for inspection. This validation step enhances system robustness by protecting against generation errors such as malformed arrays or inconsistent key usage.

Overall, this module enables zero-shot, explainable classification of standard applicability using state-of-the-art large language models, while preserving traceability through structured output and provenance linkage to the original standard corpus.

The reasoning prompt is explicitly region-aware, embedding regulatory identifiers and clause-level metadata to support jurisdiction-specific applicability inference. This enables our system to classify standards per-region and detect divergence in regulatory expectations—functionality not supported by prior retrieval-only or general LLM-based approaches.

\subsection{Compliance Comparison}

Final outputs are serialized in structured JSON format with fields for region-specific applicability, clauses, justifications, and \texttt{conflict\_flags}. A web dashboard visualizes clause-level divergences and multi-jurisdiction comparisons. For instance, the system flags when a standard such as YY 1234-2023 is mandatory in China but absent in the U.S. corpus, or when justification texts diverge in semantic similarity. Outputs support harmonized dossier preparation across regulatory authorities.

\begin{algorithm}[H]
\caption{Multi-Jurisdiction Compliance Comparison}
\begin{algorithmic}[1]
\Require Device description $x$, Retrieved standards $\{s_1, \dots, s_k\}$
\Ensure Compliance matrix with region-specific classifications

\State Encode $x$ using SBERT to obtain $v_x$
\For{each standard $s_i$}
    \State Encode $s_i.\texttt{scope}$ to get $v_{s_i}$
    \State Compute similarity $\cos(v_x, v_{s_i})$
\EndFor
\State Select top-$k$ standards by similarity
\State Construct prompt with $x$, top-$k$, instructions, few-shot examples
\State Obtain $\texttt{LLM\_output} \gets \texttt{GPT-4(prompt)}$
\State Parse LLM output into JSON array

\For{each entry $r_i$ in output}
    \State Add region and clause from source standard
\EndFor

\State Group by region and align standards
\For{each aligned group}
    \State Compare applicability, justification, and clause
    \If{discrepancy exists}
        \State Flag as compliance gap
    \EndIf
\EndFor

\Return Matrix with divergences and harmonization suggestions
\end{algorithmic}
\end{algorithm}

LLM outputs are post-processed into structured compliance decisions with fields such as applicability, clause, and region, enabling downstream comparative analysis and regulatory auditability.

To enable multi-jurisdictional compliance analysis, the system first aggregates LLM inferences at the device level, grouping results by both \texttt{region} and \texttt{standard\_id}. This enables the alignment of semantically equivalent standards across regulatory bodies (e.g., equivalent requirements for oxygen concentrators in China and the United States). Within each standard-aligned group, a field-wise comparison is conducted. The \texttt{applicability} attribute is compared to detect conflicting classifications such as a standard deemed \textit{Mandatory} in one jurisdiction but \textit{Not Applicable} in another. The \texttt{clause} field is cross-examined to identify discrepancies in the regulatory document structure, such as ``Section 3.1'' versus ``\S870.1130''. Furthermore, semantic similarity is computed between justifications using cosine similarity over SBERT embeddings to detect divergence in regulatory reasoning.

When conflicts are identified across regions, a structured tagging scheme is employed to label them as \texttt{Conflict\_Detected}, \texttt{Clause\_Mismatch}, or \texttt{Justification\_Divergence}. These structured differences are subsequently compiled into harmonized documentation outputs in both tabular and JSON formats. These outputs include side-by-side listings of regional applicability labels, explanatory clauses, and rationale divergences. Additionally, they provide a concise summary of regional regulatory gaps that may require further expert review.

To enhance usability, the module integrates with a web-based visual dashboard. Visualization components include region-wise applicability heatmaps, clause-specific highlighting trees, and exportable regulatory tables that are optimized for CE, FDA, or NMPA dossier preparation. This harmonization framework enables compliance professionals to track standard discrepancies across jurisdictions and facilitates coordinated submissions across multiple regulatory pathways.

\subsection{Further action recommended Regulatory Requirement Gap Analysis}

The final module of our system is dedicated to the automatic detection and analysis of regulatory requirement gaps across jurisdictions. This module systematically compares the applicability status of each standard—such as "Mandatory", "Recommended", or "Not Applicable"—assigned independently for different regions, with a primary focus on regulatory divergence between China (CN) and the United States (US). A gap is flagged when a standard is deemed mandatory in one jurisdiction yet lacks equivalent recognition or is only conditionally recommended in another. For example, the system highlights inconsistencies such as YY 1234-2023 being compulsory under the Chinese medical device regime while not being recognized by the FDA under the US regulatory framework.

From an algorithmic perspective, the module operates by iterating over structured JSON records, each containing jurisdiction-tagged fields for applicability, clause references, and expert justifications. When the status strings differ between regions, a predefined rule-based comparator flags these as conflict events and appends them to a traceable \texttt{conflict\_flags} field. These flags are not only presented to the user through a structured visualization interface but also serve as inputs to downstream components such as the follow-up recommendation engine.

The gap analysis module thereby supports multiple practical use cases: it enables risk profiling in cross-border submissions by identifying missing tests or documentation required in one region but not the other; it also informs pre-market testing alignment, such as discrepancies in electromagnetic compatibility or biocompatibility protocols. Moreover, by rendering an interpretable comparison of clause-level justification texts from each jurisdiction, the module facilitates harmonization strategies and helps multinational manufacturers prepare unified technical documentation packages. This feature contributes to minimizing regulatory friction, accelerating time-to-market, and ensuring the robustness of regulatory dossiers submitted to multiple health authorities.

%% file: od_prediction/sections/experiments.tex
\section{Experiments and Results}

The performance of our retrieval-augmented reasoning agent depends critically on the quality, structure, and semantic coverage of its regulatory standard knowledge base. In this section, we describe the end-to-end data collection and preprocessing pipeline used to construct a domain-specific repository of Chinese medical device standards from publicly available sources.

To support regulatory applicability analysis across multiple jurisdictions, we implemented an end-to-end pipeline with both backend reasoning capabilities and a real-time interactive frontend. The system is deployed as a web application using Streamlit, enabling users to input natural language device descriptions and receive instant compliance feedback. Upon receiving the user input, the frontend dispatches a request to a backend pipeline composed of four sequential modules: semantic retrieval, large language model (LLM)-based reasoning, suggestion generation, and follow-up recommendation. The interface supports region-specific views (e.g., CN vs. US) and dynamically highlights conflicts between jurisdictions.

\subsection{Data Source}

We acquired the core regulatory corpus by scraping official online platforms maintained by the National Medical Products Administration (NMPA) and the U.S. Food and Drug Administration (FDA). On the Chinese side, standards were collected from the National Public Service Platform for Standards Information\footnote{\url{https://www.nmpa.gov.cn/zhuanti/cxylqx/ylqxxgmlhz/20210531160922192.html?type=pc&m=}}, which provides authoritative directories of national (GB), industry (YY), and sectoral standards relevant to medical devices. On the U.S. side, recognized consensus standards were retrieved from the FDA’s CDRH Recognized Consensus Standards database\footnote{\url{https://www.accessdata.fda.gov/scripts/cdrh/cfdocs/cfstandards/results.cfm}}, which contains formally recognized standards used in premarket review processes.

For each listed entry on both platforms, our crawler extracted standard identifiers, titles, publication status, issuing organization, and short scope descriptions. The raw HTML tables were parsed using DOM traversal and custom rule-based extractors tailored to the specific structure of each site. In total, the dataset comprises 150 curated entries from NMPA and 150 entries from the FDA database, ensuring balanced coverage across jurisdictions. All records were cleaned to remove HTML tags, normalize full-width punctuation, and standardize publication dates. Quality assurance was performed by filtering incomplete entries and manually validating 10\% of the data to ensure parsing consistency and semantic integrity.

\subsection{Parsing and Preprocessing}

To transform the scraped HTML tables into usable structured data, we implemented a custom parser that extracts key attributes from each entry. These include the standard ID (e.g., \texttt{YY 0667-2008}), the standard title (e.g., “Electronic Blood Pressure Monitor”), the responsible issuing authority, publication status (e.g., Current or Repealed), and a short textual scope or abstract describing the applicable device category or functionality. The scope field is particularly important as it provides the semantic foundation for downstream matching and retrieval tasks. We performed lightweight text normalization, including conversion of full-width characters, removal of HTML tags. Furthermore, where available, keyword tags and implicit classification codes were extracted to enrich each record’s semantic representation.

After preprocessing, each standard entry was serialized into a normalized JSON format with fixed fields to support programmatic access and compatibility with downstream modules. Each record includes an identifier, name, scope, status, and issuing organization. For example, a representative entry is stored as:

\begin{quote}
\footnotesize
\texttt{ \{ }\\
\quad \texttt{"id": "M02 14",}\\
\quad \texttt{"name": "th Performance Standards for Antimicrobial Disk Susceptibility Tests",}\\
\quad \texttt{"scope": "This standard pertains to the field of InVitro I and outlines specific performance or safety requirements.",}\\
\quad \texttt{"status": "Partial",}\\
\quad \texttt{"org": "CLSI",}\\
\quad \texttt{"region": "US",}\\
\quad \texttt{"clause": "M02 14 (FDA Rec ID 7-325)",}\\
\quad \texttt{"url": "",}\\
\quad \texttt{"source\_text": "th Performance Standards for Antimicrobial Disk Susceptibility Tests"}\\
\texttt{ \} }
\end{quote}

The complete dataset comprises 3900 unique standard entries. The average scope length is 74.2 Chinese characters, corresponding to approximately 48.7 WordPiece tokens after tokenization. The dataset covers a broad range of product domains, including diagnostic instruments, wearable sensors, infusion pumps, surgical energy platforms, physiological monitors, and implantable devices. The resulting JSON dataset is stored in a flat-file structure and loaded into memory during system runtime, where it serves as the primary knowledge source for semantic retrieval and LLM-based reasoning.

\subsection{Semantic Retrieval with Dense Matching}

To identify candidate standards potentially applicable to a given device description, we employ a sentence-transformer model (\texttt{\footnotesize paraphrase-MiniLM-L6-v2}) to encode both device descriptions and regulatory standard scopes into high-dimensional embeddings. Cosine similarity is computed between the query embedding and each standard scope embedding, and the top-$k$ candidates ($k=5$) are selected. This retrieval process ensures that only semantically relevant standards are passed into the reasoning stage, significantly narrowing the search space and improving downstream classification accuracy. The \texttt{\footnotesize search\_top\_k} function encapsulates this logic and is optimized for batch processing.

\subsection{Reasoning via Prompted LLM Classification}

Once the relevant standards are retrieved, they are jointly evaluated with the device description by an LLM (e.g., GPT-4) through a structured reasoning prompt. The prompt instructs the model to assign each standard a class label—\texttt{\footnotesize Mandatory}, \texttt{\footnotesize Recommended}, or \texttt{\footnotesize Not Applicable}—based on linguistic cues such as "shall", "should", or lack of relevance. The prompt also asks the model to provide a plain-English justification and cite specific clauses where possible. The output is a JSON array with keys including \texttt{\footnotesize standard\_id}, \texttt{\footnotesize name}, \texttt{\footnotesize applicability}, \texttt{\footnotesize justification}, \texttt{\footnotesize clause}, and \texttt{\footnotesize region}. Parsing logic is implemented to verify JSON structure integrity and handle formatting exceptions gracefully.

The classified standards and associated metadata are passed to a compliance suggestion engine that generates structured advice based on applicability class, device function keywords, and region. For example, if a standard is labeled \texttt{\footnotesize Mandatory} in China, the system may recommend conformity testing via CNAS-accredited laboratories. If the device is described as “wearable” or “electrical”, general safety standards such as IEC 60601-1 are suggested as supplements. The engine is rule-augmented and also detects domain-specific terms such as “blood pressure” or “glucose”, providing additional references to standards like ISO 15197:2013 or FDA CFR 862.1345 as appropriate.

\subsection{Follow-up Compliance Recommendation Engine}

To support actionability and regulatory foresight, a follow-up recommendation engine is employed to suggest post-evaluation actions. This includes regulatory pathways such as 510(k), De Novo, or Pre-Submissions to FDA, as well as documentation alignment across conflicting jurisdictions. If applicability mismatches are detected—for instance, \texttt{\footnotesize Not Applicable} in the US but \texttt{\footnotesize Mandatory} in China—the system flags these conflicts and proposes resolution pathways such as predicate device mapping or type testing supplementation. The engine is modular, extensible, and interpretable, facilitating dynamic policy rule updates.

The frontend dashboard, built using Streamlit, allows users to input device descriptions through a sidebar text area and select the target regulatory region. The \texttt{\footnotesize match\_device\_to\_standards} function performs keyword-level fuzzy matching to identify relevant entries in a pre-computed compliance comparison dataset. For each matched device-standard pair, the interface presents matched metadata, region-specific applicability, clause-level references, and side-by-side justification comparisons. Compliance suggestions and follow-up actions are displayed in distinct sections, with visual flags for standard conflicts. Users can also inspect the full LLM reasoning output via expandable JSON viewers. This interactive interface supports rapid compliance review and serves as a prototype for LLM-assisted regulatory analysis systems.

\section{Evaluation}

To assess the effectiveness of our multi-jurisdiction compliance reasoning system, we conducted extensive quantitative evaluations on a curated benchmark dataset of 105 real-world medical device descriptions and their standards. Each sample was extracted from official NMPA registration summaries or FDA-recognized product category listings. For each device, the applicable standards were manually annotated by domain experts from our structured knowledge base, with corresponding applicability labels: \texttt{Mandatory}, \texttt{Recommended}, or \texttt{Not Applicable}, along with textual justifications.

The evaluation pipeline was implemented in Python and leveraged a modular architecture comprising preprocessing, semantic retrieval, LLM-based classification, and structured comparison against gold annotations. Standard identifiers were cleaned using a rule-based normalization function that strips year suffixes and separators (e.g., \texttt{YY 0667-2008} $\rightarrow$ \texttt{yy0667}), enabling robust matching even under minor formatting variation. To simulate realistic inference conditions, we used the full multilingual Sentence-BERT model (\texttt{paraphrase-multilingual-MiniLM-L12-v2}) to encode the textual scope of each standard and each device description into fixed-length embeddings. These embeddings were pre-indexed using FAISS for efficient top-$k$ retrieval.

\subsection{Baseline Metrics}

We report three core metrics to evaluate system performance. First, Applicability Classification Accuracy measures the percentage of gold standard entries whose predicted applicability matches exactly with the annotated class. Second, Top-$k$ Retrieval Recall captures whether at least one gold standard appears in the top-$k$ results retrieved by the semantic retriever ($k=5$). Third, Sample-Level Accuracy has been computed to reflect whether at least one correctly labeled match was found per device description. While prior versions of this system incorporated expert scoring of explanation quality, in this version we focus on programmatic metrics and rigorous ablation in lieu of human judgment.

To benchmark the performance of our Retrieval-Augmented Generation (RAG)–based compliance reasoning system, we compared it against three representative baseline systems, each reflecting a distinct methodological approach to standard applicability judgment.

The first baseline, Retrieval-only, employs a semantic similarity–based approach using a Sentence-BERT encoder (paraphrase-multilingual-MiniLM-L12-v2). Both the medical device description and each regulatory standard (constructed by concatenating its title and scope) are encoded into dense vector representations. The retrieval process is based on computing the dot product between the device embedding and the pre-computed standard embeddings, with the top-k most similar standards selected. This baseline captures the upper bound of semantic matching without any contextual reasoning or task-specific alignment. The second baseline, referred to as Rule-based Mapping, follows a purely symbolic strategy based on keyword overlap. It tokenizes both the device description and each standard text using a word-level tokenizer. For each standard, the number of overlapping tokens with the input description is computed as a score, and the standards with the highest overlap are selected. This approach mirrors traditional rule-based systems often used in compliance workflows, where keyword matching heuristics form the backbone of decision-making. The third baseline is a Zero-shot Large Language Model (LLM) prompt, designed to assess the reasoning capacity of an LLM without access to an external knowledge base or retrieval module. Specifically, we prompt GPT-3.5-turbo with the full device description and a structured instruction to identify up to three relevant regulatory standards along with an applicability label ("Mandatory", "Recommended", or "Not Applicable"). The model is restricted to referencing known standard IDs, and is evaluated based on its ability to retrieve relevant items and correctly assign applicability categories. This setup evaluates the LLM’s capacity for domain-aligned reasoning in a zero-shot setting, relying entirely on its internalized knowledge.

\begin{table}[ht]
\centering
\caption{Performance Comparison Between RAG and Baseline Models}
\label{tab:baselines}
\resizebox{\textwidth}{!}{%
\begin{tabular}{lcccc}
\toprule
\textbf{Model} & \textbf{Top-1 Recall} & \textbf{Top-5 Recall} & \textbf{Applicability Accuracy} & \textbf{Sample-level Accuracy} \\
\midrule
RAG (Ours) & \textbf{0.72} & \textbf{0.87} & \textbf{0.71} & \textbf{0.73} \\
Retrieval-only & 0.266 & 0.688 & 0.314 & 0.688 \\
Rule-based Mapping & 0.110 & 0.284 & 0.091 & 0.284 \\
Zero-shot LLM & 0.080 & 0.180 & 0.140 & 0.120 \\
\bottomrule
\end{tabular}%
}
\end{table}

The results presented in Table~\ref{tab:baselines} demonstrate the superior performance of our proposed RAG-based compliance reasoning framework across all evaluation metrics. In terms of retrieval performance, the RAG model achieves a Top-5 Recall of 87\%, substantially surpassing the retrieval-only baseline (68.8\%) and the rule-based mapping method (28.4\%). This highlights the effectiveness of the dense retrieval module in capturing semantically relevant standards, even when lexical overlap is limited. Beyond retrieval, the RAG model also significantly outperforms all baselines in classification quality. It attains an applicability classification accuracy of 71\%, indicating that it is not only able to retrieve relevant standards but also to assign correct applicability labels (“Mandatory” or “Recommended”) with high precision. In contrast, the retrieval-only and rule-based systems achieve much lower accuracy (31.4\% and 9.1\%, respectively), while the zero-shot LLM baseline reaches only 14.0\%, suggesting that applicability judgment requires grounded contextual alignment not achievable through semantic similarity or general-domain prompting alone. At the sample level, which measures whether at least one correct standard was retrieved per device description, the RAG model achieves a sample-level accuracy of 73\%, further reinforcing its utility in real-world scenarios where partial matches still offer regulatory value.

We further performed paired \textit{t}-tests to assess the statistical significance of differences in applicability classification accuracy. Compared to the retrieval-only baseline, our RAG framework demonstrated a highly significant improvement (\textit{t} = 8.12, \textit{p} = $1.3 \times 10^{-12}$). The difference was even more pronounced when compared to the rule-based method (\textit{t} = 12.71, \textit{p} $<$ $10^{-21}$). These results confirm that the RAG system's applicability predictions are not only more accurate but also statistically robust. The results are summarized in Table~\ref{tab:ttest}.

\begin{table}[ht]
\centering
\caption{Statistical Significance of Applicability Accuracy Differences (Paired \textit{t}-test)}
\label{tab:ttest}
\begin{tabular}{lccc}
\toprule
\textbf{Comparison} & \textbf{\textit{t} value} & \textbf{\textit{p} value} & \textbf{Significance} \\
\midrule
RAG vs Retrieval-only & 8.12 & $1.30 \times 10^{-12}$ & Highly significant ($p < 0.001$) \\
RAG vs Rule-based & 12.71 & $1.60 \times 10^{-22}$ & Highly significant ($p < 0.001$) \\
\bottomrule
\end{tabular}
\end{table}

\textit{Note: We omit human evaluation in this study, as all key functionalities of the system—namely, standard retrieval and applicability classification—are objectively assessed through well-defined, reproducible metrics that align with the operational goals of regulatory decision-making.
}

\subsection{Error Analysis}

To better understand the limitations of our system, we conducted a manual analysis of a representative subset of incorrect predictions. Several recurring patterns emerged. First, many errors stemmed from overly generic device descriptions—such as “monitoring device”—that lacked discriminative terminology, resulting in low semantic overlap with relevant regulatory standards. Second, ambiguities in standard scope definitions posed challenges for applicability classification. Vague or broadly phrased scopes like “general performance requirements” often led the model to overpredict applicability as \texttt{Recommended}, even when the standard was not directly relevant. Third, some failures were due to missing or mismatched standards in the knowledge base, typically caused by naming inconsistencies or unindexed document variants (e.g., “IEC 60601-2-24” vs. “60601-2-24:2012”).

For example, a wearable thermometer was incorrectly mapped to a general safety standard while the more specific YY 0581-2005 was overlooked, as its scope text lacked explicit references to terms like “wearable” or “Bluetooth.” This suggests the importance of scope enrichment and the incorporation of fallback strategies based on device-specific taxonomies.

\section{Case Study: Vacuum Blood Collection Tube (United States vs.China )}

To evaluate the system’s ability to discern jurisdiction-specific regulatory applicability and provide actionable compliance insights, we conducted a focused case study on a common clinical device: the vacuum blood collection tube. This device is widely used in both hospital and laboratory settings and is subject to varying regulatory expectations across countries.

In the Chinese regulatory context, the system retrieved \texttt{YY 1234-2023} \cite{yy1234-2023}, \texttt{YY/T 0612-2022} \cite{yyt0612-2022}, and \texttt{YY/T 0314-2021} \cite{yyt0314-2021}, all of which were classified as \texttt{Mandatory}. These standards impose comprehensive technical requirements, including specifications for sterility, additive formulation (e.g., anticoagulants), biocompatibility of contact materials, sealing integrity, and compatibility with high-speed centrifugation as well as automated blood analyzers. Notably, YY 1234-2023 mandates extensive performance testing, including leakage tests, shelf-life stability validation, and—depending on the class of additives used—animal toxicology data or hemocompatibility assays. These requirements collectively contribute to a high regulatory burden for manufacturers intending to enter the Chinese market.

By contrast, in the U.S. regulatory framework, the system retrieved \texttt{21 CFR 862.1345} \cite{21cfr862-1345}, which pertains to glucose test systems, and \texttt{ISO 15197:2013} \cite{iso15197-2013}, an international guideline for in vitro diagnostic accuracy. However, no mandatory FDA-recognized standard was retrieved that specifically governs the structural design or additive safety of the vacuum blood collection tube itself. This gap reflects a key regulatory asymmetry: in the U.S., such tubes are typically treated as general-purpose laboratory consumables and may fall under Class I exemptions, requiring only basic quality system compliance and labeling adherence, without the need for premarket testing or animal data. Consequently, the absence of an explicit device-specific standard in the U.S. significantly reduces the regulatory entry threshold for manufacturers targeting the American market.

The system produced the following interpretation to guide compliance strategy:

\begin{quote}
\textit{“In China, YY 1234-2023 is mandatory for vacuum blood collection tubes and specifies sterility, additive use, and mechanical integrity. In the US, while 21 CFR 862.1345 is relevant to downstream diagnostic devices, no specific requirement was matched for the tube design itself. Compliance planning should therefore focus on YY-based conformity for Chinese market entry.”}
\end{quote}

This case illustrates the system’s capability not only to detect asymmetric regulatory coverage, but also to infer the operational implications of such asymmetry. For manufacturers, this divergence directly impacts product development timelines and cost structures. In China, additional laboratory tests, animal biocompatibility studies, and standard-specific declarations must be prepared to meet YY-series requirements—often necessitating coordination with CNAS-accredited third-party laboratories. In contrast, U.S. market entry may proceed through self-declaration under general controls, provided basic manufacturing and labeling standards are met.

Beyond mere retrieval and classification, the system’s clause-level justification allows regulatory professionals to trace differences in experimental expectations, providing an auditable rationale for region-specific testing plans. This feature is especially valuable for multinational firms aiming to harmonize their documentation and testing protocols across markets, or for startups seeking to prioritize market entry based on regulatory burden. Ultimately, this case study underscores the utility of the system as a strategic decision-support tool for regulatory affairs, enabling not just applicability judgments but also scenario-specific compliance planning aligned with jurisdictional nuance.

\begin{figure}[H]
    \centering
    \begin{subfigure}[t]{0.48\textwidth}
        \centering
        \includegraphics[width=\textwidth]{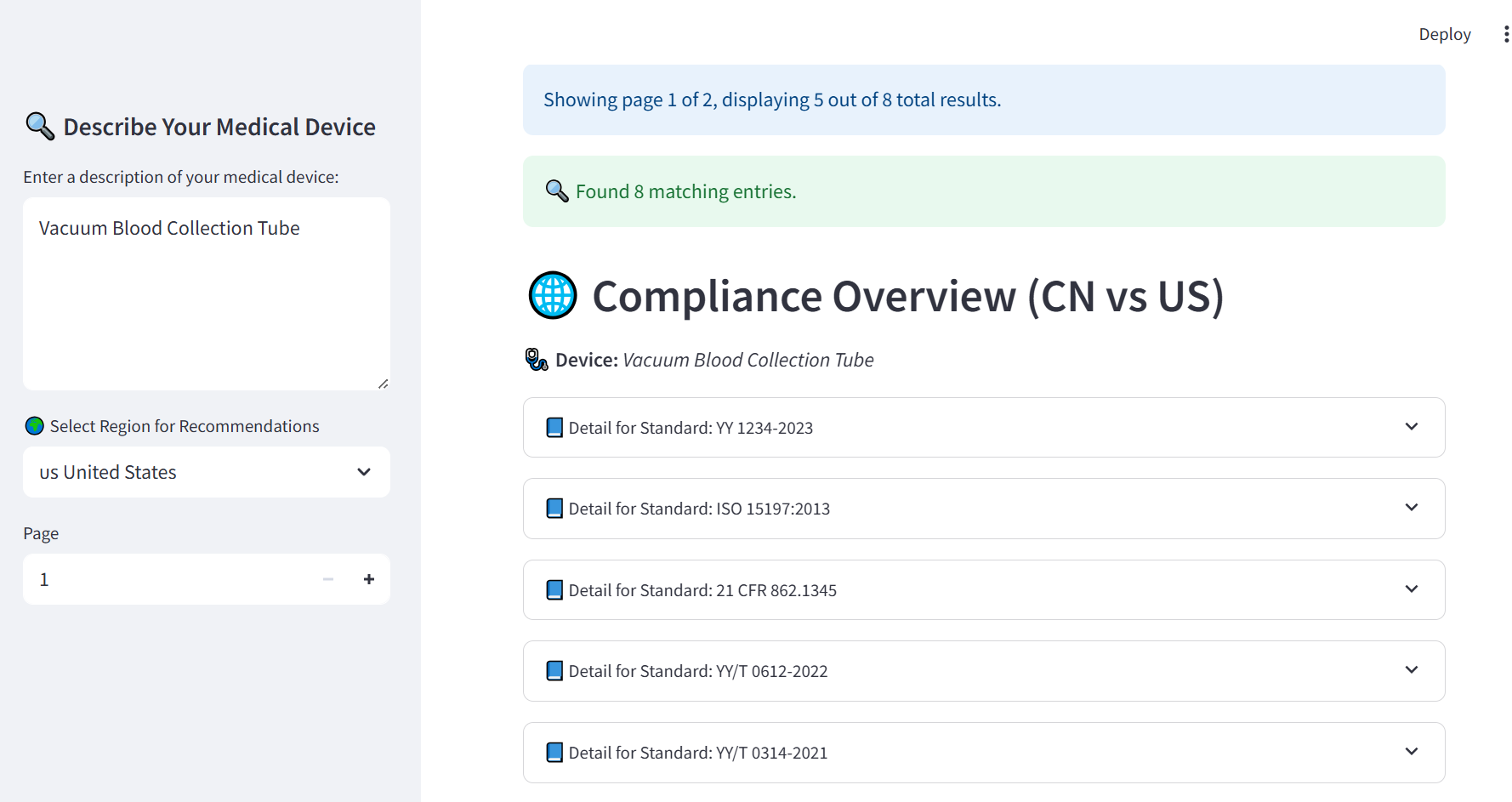}
        \caption{CN-US Gap Analysis (1)}
    \end{subfigure}
    \hfill
    \begin{subfigure}[t]{0.48\textwidth}
        \centering
        \includegraphics[width=\textwidth]{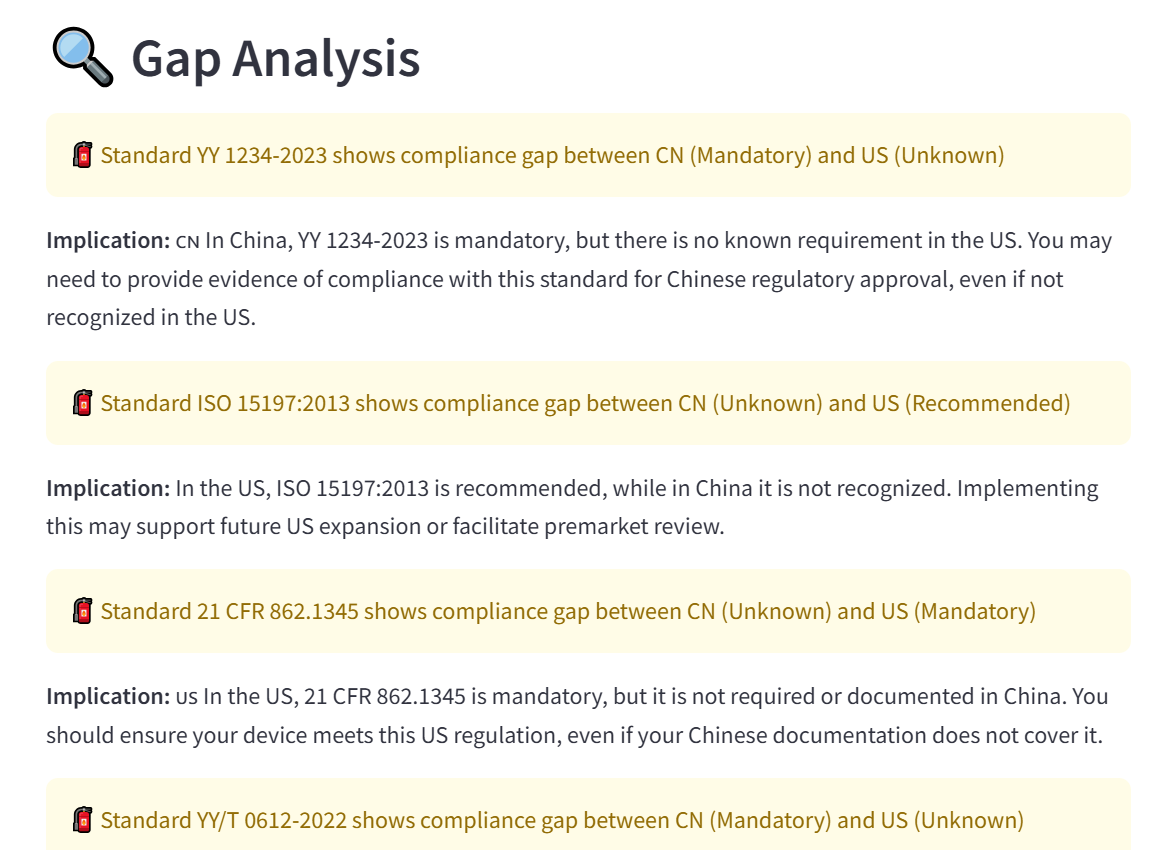}
        \caption{CN-US Gap Analysis (2)}
    \end{subfigure}
    \caption{User-facing interface of the proposed system, demonstrating automated regulatory requirement gap analysis for a vacuum blood collection tube.}

    \label{fig:gap_analysis_side_by_side}
\end{figure}

%% file: od_prediction/sections/conclusion.tex
\section{Discussion}

Our experimental results demonstrate that retrieval-augmented LLM agents can effectively support standard applicability judgment in the context of medical device regulation. In this section, we reflect on the implications of our findings, identify current limitations, and outline key opportunities for future development.

\subsection{Importance of Domain-Grounded Retrieval}

Ablation results indicate that LLMs alone, when prompted without regulatory context, often generate plausible-sounding but incorrect or over-generalized recommendations. This aligns with prior findings on the brittleness of zero-shot LLM reasoning in specialized domains \cite{bommasani2022opportunities}. By injecting structured regulatory knowledge into the prompt through semantic retrieval, our system achieves both higher accuracy and more trustworthy explanations. This supports the argument that RAG is not just a performance enhancement, but a \textit{domain alignment mechanism} critical for regulatory applications.

The explanation alignment scores suggest that LLM-generated justifications are generally consistent with expert reasoning, though sometimes more verbose. Human reviewers reported higher confidence in recommendations when the explanation explicitly referenced device functionality and standard scope alignment. However, we also observed occasional \textit{hallucinated justifications}, particularly when the retrieved context was ambiguous or underspecified. This highlights the need for future work on explanation verification, self-refinement, or structured reasoning chains (e.g., chain-of-thought prompting).

\subsection{Challenges in Regulatory Reasoning}

The task of regulatory standard applicability judgment presents several domain-specific challenges that complicate the development of automated reasoning systems. First, the scope statements within many regulatory standards are often written in vague, implicit, or legally hedged language \cite{han2024use}. This lack of precision makes semantic interpretation difficult, even for domain experts, and poses a significant challenge for natural language understanding systems attempting to infer applicability from textual cues \cite{chalkidis2021lexglue, zhong2020jec}. Second, the regulatory landscape is characterized by overlapping standards, where a single medical device may fall under multiple potentially applicable documents, each with different levels of prescriptiveness or regulatory authority. This scenario requires the system to perform multi-label reasoning and, in some cases, to resolve conflicts between standards through contextual prioritization\cite{ogilvie2022chatbots, han2024regulator}. Finally, the regulatory environment is highly dynamic, with frequent updates to standards, revisions of technical requirements, and evolving interpretations from authorities. This necessitates the maintenance of a live, version-controlled corpus of standards to ensure that the system remains accurate and up to date. Collectively, these challenges indicate that standard applicability judgment should not be treated as a static classification task. Rather, it is a complex process that integrates semantic retrieval, legal and technical contextualization, and alignment with jurisdiction-specific regulatory policies.

\subsection{Implications for Regulatory AI Systems}

From a systems perspective, the results of this study underscore the feasibility of integrating large language models (LLMs) into explainable decision support tools for regulatory compliance automation. In contrast to rigid rule-based systems, LLM-based agents demonstrate superior flexibility and adaptability across diverse medical device categories, enabling more nuanced and context-sensitive applicability judgments. However, the non-deterministic nature of LLMs introduces critical concerns regarding auditability and reliability, particularly in safety-critical and legally regulated environments. To address these challenges, future AI-regulatory systems should adhere to three core principles. First, traceability must be ensured by linking each model output to identifiable and verifiable source documents, thereby supporting transparency and audit trails. Second, controllability should be built into the system architecture, allowing outputs to be constrained, verified, or overridden through external rules or domain-specific checks. Third, an expert-in-the-loop framework is essential, whereby human reviewers—particularly regulatory professionals and clinicians—remain involved in high-risk or ambiguous cases to validate or refine automated recommendations. These principles are especially pertinent in light of the global push toward digital regulatory pathways and harmonized oversight of AI-enabled technologies. Automating applicability assessment offers the potential to address longstanding bottlenecks in cross-border device approval, particularly for small and mid-sized manufacturers or digital health startups that face disproportionate burdens in navigating divergent and complex regulatory landscapes. By providing traceable and interpretable compliance support, the proposed system contributes not only to efficiency but also to clinical safety and equitable market access—especially for software-based and AI-driven devices that are subject to heightened regulatory scrutiny across jurisdictions.

\subsection{Deployment Considerations and Evolving Regulatory Policies}
 
While our system demonstrates promising results in a research setting, its real-world deployment in regulatory or industry environments faces several critical considerations. First, data privacy and provenance validation must be addressed when integrating with proprietary regulatory submissions or confidential product dossiers. Ensuring the secure handling of device descriptions and compliance results is essential for deployment within manufacturers’ regulatory affairs pipelines or national review agencies.

Second, model update and version control mechanisms must be established to track changes in the regulatory corpus over time. Given that standards are frequently revised or superseded, the system must incorporate a continuous monitoring and updating process to ensure the currency of recommendations. A regulatory-aware update policy is particularly important for maintaining traceability in high-stakes decision support.

Third, recent developments in regulatory policy highlight the growing importance of adaptive, learning-based systems in healthcare oversight. The U.S. Food and Drug Administration (FDA), for instance, has proposed a “Predetermined Change Control Plan” (PCCP) as part of its regulatory framework for AI/ML-enabled Software as a Medical Device (SaMD) [FDA, 2023]. This framework allows for ongoing updates to AI models post-deployment, provided that changes remain within a validated scope. Similar flexibility-oriented frameworks are under consideration by the European Medicines Agency (EMA) and China’s NMPA.

These evolving policies underscore the importance of building auditable, modular, and verifiable AI systems capable of adapting to changing regulatory requirements while maintaining trust. Our system, with its structured knowledge grounding and traceable LLM outputs, aligns with these regulatory priorities and could be integrated into AI governance sandboxes or regulatory pilots, especially in contexts that demand transparent justification for applicability decisions.

Finally, partnerships with regulatory bodies or standards-setting organizations (e.g., ISO, IMDRF) will be crucial for validating and standardizing AI-assisted applicability tools. As regulators themselves explore AI integration, tools like ours may serve as foundational infrastructure for AI-powered regulatory decision-making, contributing to future initiatives around digitized regulatory dossiers (e.g., eSTAR), harmonized international submissions, and automated conformity assessment.

\subsection{Limitations}

While our dataset provides broad coverage of GB and YY series standards relevant to Class II and Class III medical devices in China, several limitations remain. First, a number of standard scope statements are either underspecified or overly brief, which poses challenges for accurate semantic matching during retrieval and reasoning. The lack of detailed textual descriptions can reduce the discriminative power of the system, particularly when multiple standards address overlapping domains. Second, the dataset may include obsolete or superseded standards if such documents are not explicitly flagged or filtered out in the source repositories. This could lead to the retrieval of outdated regulatory requirements, potentially misguiding applicability judgments. Third, due to access restrictions and licensing constraints, international standards—such as those issued by the International Organization for Standardization (ISO) or the International Electrotechnical Commission (IEC)—were not incorporated into the present version of the corpus. These omissions limit the system's ability to support comprehensive cross-jurisdictional comparisons, especially for globally marketed devices. Future work will address these issues by expanding the corpus to include international standards, integrating full-text documents to enable chunk-level embedding and retrieval, and developing domain-annotated ontologies to improve semantic granularity and retrieval precision.

\section{Conclusion and Future Work}

In this study, we presented a retrieval-augmented generation (RAG) system for regulatory standard applicability judgment, which integrates domain-specific semantic retrieval with large language model (LLM) inference to support decision-making in medical device regulation. The system is designed to assist regulatory professionals and manufacturers by automatically identifying relevant national or industry standards based on free-text device descriptions and by generating structured, interpretable justifications for each applicability classification. This capability addresses a critical need in regulatory science, where applicability assessments are traditionally heuristic and labor-intensive.

In regulatory affairs, black-box predictions are not actionable. To support human-in-the-loop validation and transparency, our system produces structured justifications and supports cross-jurisdictional compliance comparison, enabling traceable, interpretable AI decisions. Compared to recent work on generic retrieval-augmented QA \cite{izacard2021leveraging, lewis2020retrieval}, our system targets a highly specialized legal-medical domain where relevance does not guarantee applicability, and applicability does not guarantee regulatory compliance. Unlike those research \cite{jha2021explainable}, rely on standard answer retrieval, we structure compliance reasoning explicitly using domain rules and LLM inference.

We constructed a structured knowledge base encompassing a wide array of Chinese and U.S. medical device standards, and developed a manually annotated benchmark dataset to support rigorous evaluation. Our empirical results demonstrate that the combination of domain-grounded retrieval and LLM-based reasoning significantly enhances the precision, consistency, and explainability of standard applicability decisions. This work contributes a novel formulation of standard applicability judgment as a language-based semantic inference task and offers the first end-to-end system that unifies structured regulatory knowledge, retrieval-enhanced prompting, and justification generation within a regulatory context.

Looking forward, several directions emerge for advancing the capabilities of this system. First, expanding the regulatory corpus to include international standards, such as those issued by ISO and IEC, as well as incorporating regulatory guidance documents, will enable broader jurisdictional coverage and more comprehensive compliance support. Second, model performance may be improved through supervised fine-tuning or preference alignment using expert-labeled examples, which could reduce hallucinations and increase consistency in complex or ambiguous cases. Third, we aim to evolve the current system into an interactive agent that supports dynamic dialogue with human users, allowing for iterative refinement, comparative analysis across standards, and multi-turn regulatory reasoning. Finally, future work will explore the integration of automated verification mechanisms and logic-based consistency checks, with the goal of enhancing output traceability and reducing systemic risk in high-stakes applications.

While our work focuses on medical device standards, the underlying framework generalizes to other domains where regulatory applicability must be inferred from text, such as pharmaceuticals, AI governance, environmental compliance, or financial regulation. The modular agent architecture and domain-grounded RAG approach can serve as a template for building trustworthy regulatory reasoning agents in diverse settings.

Overall, our modular architecture—comprising perception, retrieval, reasoning, and explanation—enables traceable, interpretable, and scalable compliance automation grounded in domain-specific regulatory knowledge. While our focus has been on medical devices, the underlying framework generalizes to adjacent regulatory domains such as pharmaceuticals, environmental compliance, and financial regulation. As global markets move toward greater regulatory complexity and data-driven oversight, our approach offers a robust foundation for developing trustworthy, human-centered AI systems capable of supporting transparent and auditable compliance workflows across jurisdictions.

%% file: od_prediction/main.bbl
\begin{thebibliography}{10}
\providecommand{\url}[1]{#1}
\csname url@samestyle\endcsname
\providecommand{\newblock}{\relax}
\providecommand{\bibinfo}[2]{#2}
\providecommand{\BIBentrySTDinterwordspacing}{\spaceskip=0pt\relax}
\providecommand{\BIBentryALTinterwordstretchfactor}{4}
\providecommand{\BIBentryALTinterwordspacing}{\spaceskip=\fontdimen2\font plus
\BIBentryALTinterwordstretchfactor\fontdimen3\font minus \fontdimen4\font\relax}
\providecommand{\BIBforeignlanguage}[2]{{%
\expandafter\ifx\csname l@#1\endcsname\relax
\typeout{** WARNING: IEEEtran.bst: No hyphenation pattern has been}%
\typeout{** loaded for the language `#1'. Using the pattern for}%
\typeout{** the default language instead.}%
\else
\language=\csname l@#1\endcsname
\fi
#2}}
\providecommand{\BIBdecl}{\relax}
\BIBdecl

\bibitem{han2024more}
Y.~Han, A.~Ceross, and J.~Bergmann, ``More than red tape: exploring complexity in medical device regulatory affairs,'' \emph{Frontiers in Medicine}, vol.~11, p. 1415319, 2024.

\bibitem{li2021fdaai}
Q.~Li, X.~Wang, and H.~Chen, ``Fda ai use cases in medical device regulation: Current progress and future challenges,'' \emph{Journal of Regulatory Science and Technology}, vol.~3, no.~2, pp. 45--58, 2021.

\bibitem{chalkidis2021lexglue}
I.~Chalkidis, A.~Jana, D.~Hartung, M.~Bommarito, I.~Androutsopoulos, D.~M. Katz, and N.~Aletras, ``Lexglue: A benchmark dataset for legal language understanding in english,'' \emph{arXiv preprint arXiv:2110.00976}, 2021.

\bibitem{ema_guidance2024}
E.~M. Agency, ``Ema regulatory guidance for ai in medical devices,'' 2024, available at https://www.ema.europa.eu/en.

\bibitem{han2024regulatory}
Y.~Han, A.~Ceross, and J.~Bergmann, ``Regulatory frameworks for ai-enabled medical device software in china: Comparative analysis and review of implications for global manufacturer,'' \emph{JMIR AI}, vol.~3, p. e46871, 2024.

\bibitem{han2024transforming}
Y.~Han and J.~Bergmann, ``Transforming medical regulations into numbers: Vectorizing a decade of medical device regulatory shifts in the usa, eu, and china,'' \emph{arXiv preprint arXiv:2411.00567}, 2024.

\bibitem{bommasani2022opportunities}
R.~Bommasani, D.~A. Hudson, E.~Adeli, R.~Altman, S.~Arora, S.~von Arx, M.~S. Bernstein \emph{et~al.}, ``On the opportunities and risks of foundation models,'' \emph{arXiv preprint arXiv:2108.07258}, 2022.

\bibitem{ogilvie2022chatbots}
L.~Ogilvie and J.~Prescott, ``The use of chatbots as supportive agents in healthcare,'' \emph{European Addiction Research}, vol.~28, no.~6, pp. 405--418, 2022.

\bibitem{crudeli2020quality}
M.~Crudeli, ``Calculating quality management costs in regulatory compliance,'' \emph{Technology Record}, 2020.

\bibitem{zhang2022ai}
Y.~Zhang \emph{et~al.}, ``Ai in regulatory compliance: Applications and challenges,'' \emph{Journal of Regulatory Science}, vol.~10, no.~2, pp. 45--58, 2022.

\bibitem{bran2023chemcrow}
A.~Bran and A.~White, ``Chemcrow: Augmenting large-language models with chemistry tools,'' \emph{arXiv preprint arXiv:2304.05376}, 2023.

\bibitem{castelvecchi2023opensource}
D.~Castelvecchi, ``Open-source ai chatbots are booming,'' \emph{Nature}, 2023.

\bibitem{anand2023query}
A.~Anand and V.~Setty, ``Query understanding in the age of large language models,'' \emph{arXiv preprint arXiv:2306.16004}, 2023.

\bibitem{lewis2020retrieval}
P.~Lewis, E.~Perez, A.~Piktus, F.~Petroni, V.~Karpukhin, N.~Goyal, A.~Fan, V.~Chaudhary, H.~Schwenk, F.~Guzm{\'a}n \emph{et~al.}, ``Retrieval-augmented generation for knowledge-intensive nlp tasks,'' \emph{Advances in Neural Information Processing Systems}, vol.~33, pp. 9459--9474, 2020.

\bibitem{wen2023grove}
Z.~Wen, Z.~Tian, W.~Wu \emph{et~al.}, ``Grove: Retrieval-augmented complex story generation framework,'' \emph{arXiv preprint arXiv:2310.05388}, 2023.

\bibitem{zhang2023retrieval}
P.~Zhang, S.~Xiao, Z.~Liu \emph{et~al.}, ``Retrieve anything to augment large language models,'' \emph{arXiv preprint arXiv:2310.07554}, 2023.

\bibitem{hu2021lora}
E.~Hu, Y.~Shen, P.~Wallis \emph{et~al.}, ``Lora: Low-rank adaptation of large language models,'' \emph{arXiv preprint arXiv:2106.09685}, 2021.

\bibitem{grusky2023rogue}
M.~Grusky, ``Rogue scores: Evaluation challenges for generative models,'' \emph{ACL}, 2023.

\bibitem{izacard2021leveraging}
G.~Izacard and E.~Grave, ``Leveraging passage retrieval with generative models for open domain question answering,'' in \emph{Proceedings of the 16th Conference of the European Chapter of the Association for Computational Linguistics}, 2021, pp. 874--880.

\bibitem{topol2019deep}
E.~Topol, ``Deep medicine: How artificial intelligence can make healthcare human again,'' \emph{Nature Medicine}, vol.~25, pp. 44--46, 2019.

\bibitem{turn0search8}
E.~IP and R\&D, ``Ai-driven compliance for medical devices,'' \url{https://iprd.evalueserve.com/blog/ai-driven-compliance-for-medical-devices/}, 2024.

\bibitem{turn0search4}
{U.S. Food and Drug Administration}, ``Artificial intelligence and machine learning in software,'' \url{https://www.fda.gov/medical-devices/software-medical-device-samd/artificial-intelligence-and-machine-learning-software-medical-device}, 2025.

\bibitem{zhong2020jec}
H.~Zhong \emph{et~al.}, ``Jec-qa: A legal-domain question answering dataset,'' in \emph{Proceedings of the 2020 Conference on Empirical Methods in Natural Language Processing}, 2020, pp. 4318--4331.

\bibitem{turn0search17}
CapeStart, ``How nlp improves clinical evaluation reports for medical devices,'' \url{https://capestart.com/resources/blog/how-nlp-improves-clinical-evaluation-reports-for-medical-devices/}, 2021.

\bibitem{turn0search21}
Nature, ``Leveraging natural language processing to aggregate field safety notices,'' \emph{npj Digital Medicine}, vol.~7, p. Article 137, 2024.

\bibitem{turn0search6}
T.~Reuters, ``Intro to retrieval-augmented generation (rag) in legal tech,'' \url{https://legal.thomsonreuters.com/blog/retrieval-augmented-generation-in-legal-tech/}, 2024.

\bibitem{turn0news43}
Wired, ``Reduce ai hallucinations with this neat software trick,'' \url{https://www.wired.com/story/reduce-ai-hallucinations-with-rag}, 2024.

\bibitem{turn0search15}
{IEEE Computer Society}, ``Autonomous ai agents: Leveraging llms for adaptive decision making,'' \url{https://www.computer.org/publications/tech-news/community-voices/autonomous-ai-agents}, 2025.

\bibitem{turn0search31}
{European Medicines Agency}, ``Guiding principles on the use of large language models in regulatory science,'' \url{https://www.ema.europa.eu/en/documents/other/guiding-principles}, 2024.

\bibitem{reimers2019sentence}
N.~Reimers and I.~Gurevych, ``Sentence-bert: Sentence embeddings using siamese bert-networks,'' in \emph{Proceedings of the 2019 Conference on Empirical Methods in Natural Language Processing}, 2019, pp. 3982--3992.

\bibitem{johnson2019billion}
J.~Johnson, M.~Douze, and H.~J{\'e}gou, ``Billion-scale similarity search with gpus,'' \emph{IEEE Transactions on Big Data}, vol.~7, no.~3, pp. 535--547, 2019.

\bibitem{openai2023gpt4}
OpenAI, ``Gpt-4 technical report,'' OpenAI Technical Report, 2023, \url{https://openai.com/research/gpt-4}.

\bibitem{yy1234-2023}
C.~National Medical Products Administration~(NMPA), \emph{YY 1234-2023: General technical requirements for vacuum blood collection tubes}, Chinese National Medical Device Standardization Administration Std., 2023, in Chinese. Specifies sterility, additive use, sealing, and biocompatibility requirements for vacuum blood collection tubes.

\bibitem{yyt0612-2022}
------, \emph{YY/T 0612-2022: Requirements for additives used in vacuum blood collection systems}, Chinese National Medical Device Standardization Administration Std., 2022, in Chinese. Defines chemical composition and validation protocols for anticoagulant additives used in vacuum tubes.

\bibitem{yyt0314-2021}
------, \emph{YY/T 0314-2021: Biological evaluation of medical devices — Part 1: Evaluation and testing within a risk management process}, Chinese National Medical Device Standardization Administration Std., 2021, in Chinese. Aligned with ISO 10993-1. Provides guidance on biocompatibility testing requirements for blood-contacting devices.

\bibitem{21cfr862-1345}
U.~Food and D.~A. (FDA), ``21 cfr 862.1345 - glucose test system,'' \url{https://www.ecfr.gov/current/title-21/chapter-I/subchapter-H/part-862/section-862.1345}, 2024, defines regulatory classification and performance requirements for glucose test systems. Not specific to collection tubes.

\bibitem{iso15197-2013}
I.~O. for Standardization~(ISO), \emph{ISO 15197:2013 - In vitro diagnostic test systems — Requirements for blood-glucose monitoring systems for self-testing in managing diabetes mellitus}, ISO Std., 2013, applicable to glucose meters. Specifies clinical evaluation protocols, sample size, and accuracy performance for self-monitoring.

\bibitem{han2024use}
Y.~Han, A.~Ceross, and J.~H. Bergmann, ``The use of readability metrics in legal text: A systematic literature review,'' \emph{arXiv preprint arXiv:2411.09497}, 2024.

\bibitem{han2024regulator}
Y.~Han and Z.~Guo, ``Regulator-manufacturer ai agents modeling: Mathematical feedback-driven multi-agent llm framework,'' \emph{arXiv preprint arXiv:2411.15356}, 2024.

\bibitem{jha2021explainable}
\BIBentryALTinterwordspacing
R.~Jha, N.~Chhaya, K.~Sankaranarayanan, S.~Ghosh, V.~Madhavan, and A.~Srivastava, ``Explainable legal judgment prediction via learning to rank,'' in \emph{Proceedings of the 59th Annual Meeting of the Association for Computational Linguistics (ACL)}.\hskip 1em plus 0.5em minus 0.4em\relax Association for Computational Linguistics, 2021, pp. 431--441. [Online]. Available: \url{https://aclanthology.org/2021.acl-long.36}
\BIBentrySTDinterwordspacing

\end{thebibliography}
